%% file: arxiv.tex
\documentclass[letterpaper, 10 pt, journal, twoside]{IEEEtran}
\usepackage{booktabs}
\usepackage{multirow}
\usepackage{xcolor}
\usepackage{caption}
\usepackage{gensymb}
\usepackage{url}
\usepackage{epsfig} 
\usepackage{mathptmx} 
\usepackage{times} 
\usepackage{amsmath} 
\usepackage{amssymb}  
\usepackage{subfigure}
\usepackage{array}
\usepackage{color, soul}
\usepackage[colorlinks,linkcolor=red,anchorcolor=blue,citecolor=green]{hyperref}
\usepackage{colortbl}
\usepackage{tcolorbox}
\usepackage{color,xcolor}

\usepackage{cite}
\usepackage{xcolor}
\usepackage{algpseudocode}
\usepackage[export]{adjustbox}
\usepackage{balance}
\usepackage{rotating}
\usepackage{siunitx} 
\usepackage{amsfonts} 
\usepackage{textcomp}
\usepackage{stfloats}
\usepackage{verbatim}
\usepackage{indentfirst}
\usepackage{fontenc}
\usepackage{setspace}
\usepackage{lineno}
\usepackage{float}
\usepackage{makecell}
\usepackage{pifont}
\usepackage{bbding}
\usepackage{threeparttable}
\usepackage{tabularx}
\definecolor{myblue}{RGB}{27,183,251}
\definecolor{mygray}{gray}{.6}
\definecolor{mygreen1}{RGB}{81,150,111}
\definecolor{mygreen2}{RGB}{93,174,86}
\definecolor{myred}{RGB}{160,0,0}
\definecolor{myyellow}{RGB}{227,207,87}
\usepackage{mathrsfs}

\DeclareMathAlphabet{\mathcal}{OMS}{cmsy}{m}{n}
\usepackage{graphicx}

\usepackage[utf8]{inputenc}
\IEEEoverridecommandlockouts                              
\title{\LARGE \bf
Beyond Frame-wise Tracking: A Trajectory-based Paradigm \\ for Efficient Point Cloud Tracking}

\author{BaiChen Fan$^{1\dagger}$, Yuanxi Cui$^{2\dagger}$, Jian Li$^{1}$, Qin Wang$^{1}$, Shibo Zhao$^{3}$, Muqing Cao$^{3}$, Sifan Zhou$^{3*\ddagger}$
\thanks{*Corresponding Author. $\dagger$Equal Contribution.  $\ddagger$Project Leader.}
\thanks{$^{1}$BaiChen Fan, Jian Li, and Qin Wang are with the Institute of Quantum Information and Technology, Nanjing University of Posts and Telecommunications, Nanjing, China.} 
\thanks{$^{2}$Yuanxi Cui is with the Shanghai Jiao Tong University, Shanghai, China.}
\thanks{$^{3}$Sifan Zhou, Shibo Zhao and Muqing Cao are with the Robotics Institute, Carnegie Mellon University, PA, USA.}
}

\begin{document}
\maketitle
\begin{abstract}
LiDAR-based 3D single object tracking (3D SOT) is a critical task in robotics and autonomous systems. Existing methods typically follow frame-wise motion estimation or a sequence-based paradigm. However, the two-frame methods are efficient but lack long-term temporal context, making them vulnerable in sparse or occluded scenes, while sequence-based methods that process multiple point clouds gain robustness at a significant computational cost. To resolve this dilemma, we propose a novel trajectory-based paradigm and its instantiation, \textbf{TrajTrack}. TrajTrack is a lightweight framework that enhances a base two-frame tracker by implicitly learning motion continuity from historical bounding box trajectories alone—without requiring additional, costly point cloud inputs. It first generates a fast, explicit motion proposal and then uses an implicit motion modeling module to predict the future trajectory, which in turn refines and corrects the initial proposal. Extensive experiments on the large-scale NuScenes benchmark show that TrajTrack achieves new state-of-the-art performance, dramatically improving tracking precision by 3.02\% over a strong baseline while running at 55 FPS. Besides, we also demonstrate the strong generalizability of TrajTrack across different base trackers. Code is available at \textcolor{red}{\url{https://github.com/FiBonaCci225/TrajTrack}}.

\end{abstract}

\input{text/1_intro}
\input{text/2_related}
\input{text/3_method}
\input{text/4_exp}
\input{text/5_conclusion}

\bibliographystyle{ieeetr}
\bibliography{refenrences}
\end{document}

%% file: text/1_intro.tex
\vspace{-2mm}
\section{INTRODUCTION}  
Environments perception is important for autonomous driving and mobile robots, as it provides essential contextual information for scene understanding and decision-making~\cite{zhao2026advances,zhao2024balf,lu2025yo,yan2025turboreg,ma2025reinforcement,liu2025pipe,li2024mlp,ma2025energy,lu2023tf,lu2024mace,sun2024gsrender,zhao2025tartan,shi2025rethinking, wang2025target, wang2024scantd}. Within this context, 3D single object tracking (3D SOT) has emerged as a key task in computer vision, enabling fine-grained localization and continuous monitoring of individual objects—capabilities that are crucial for real-world applications such as autonomous navigation and robotic interaction~\cite{javed2022visual,cui2019point,zhou2025comptrack}. It aims to accurately track a target across frames using LiDAR or cameras. Unlike Multi-object tracking (MOT)~\cite{AB3DSOT,simpletrack,mutr3d,ada}, which is mostly based on the tracking-by-detection paradigm and highly sensitive to missed detections or identity switching, SOT only requires initial target annotation and can perform end-to-end tracking of the target without a detector, providing higher robustness and making it particularly suitable for robotic scenarios that require continuous tracking of a specified target. Compared to camera images, LiDAR point clouds offer notable advantages: robustness to light changes and the ability to capture rich geometries from environments. These strengths make it well-suited for robust perception in robotics~\cite{zhou2023fastpillars,zhou2025pillarhist}. However, challenges from point clouds' inherent sparsity, especially under occlusion, and appearance changes due to motion or blur often lead to incomplete observations, making reliable tracking difficult.

\begin{figure}[t]
    \centering 
    \includegraphics[width=\linewidth, keepaspectratio]{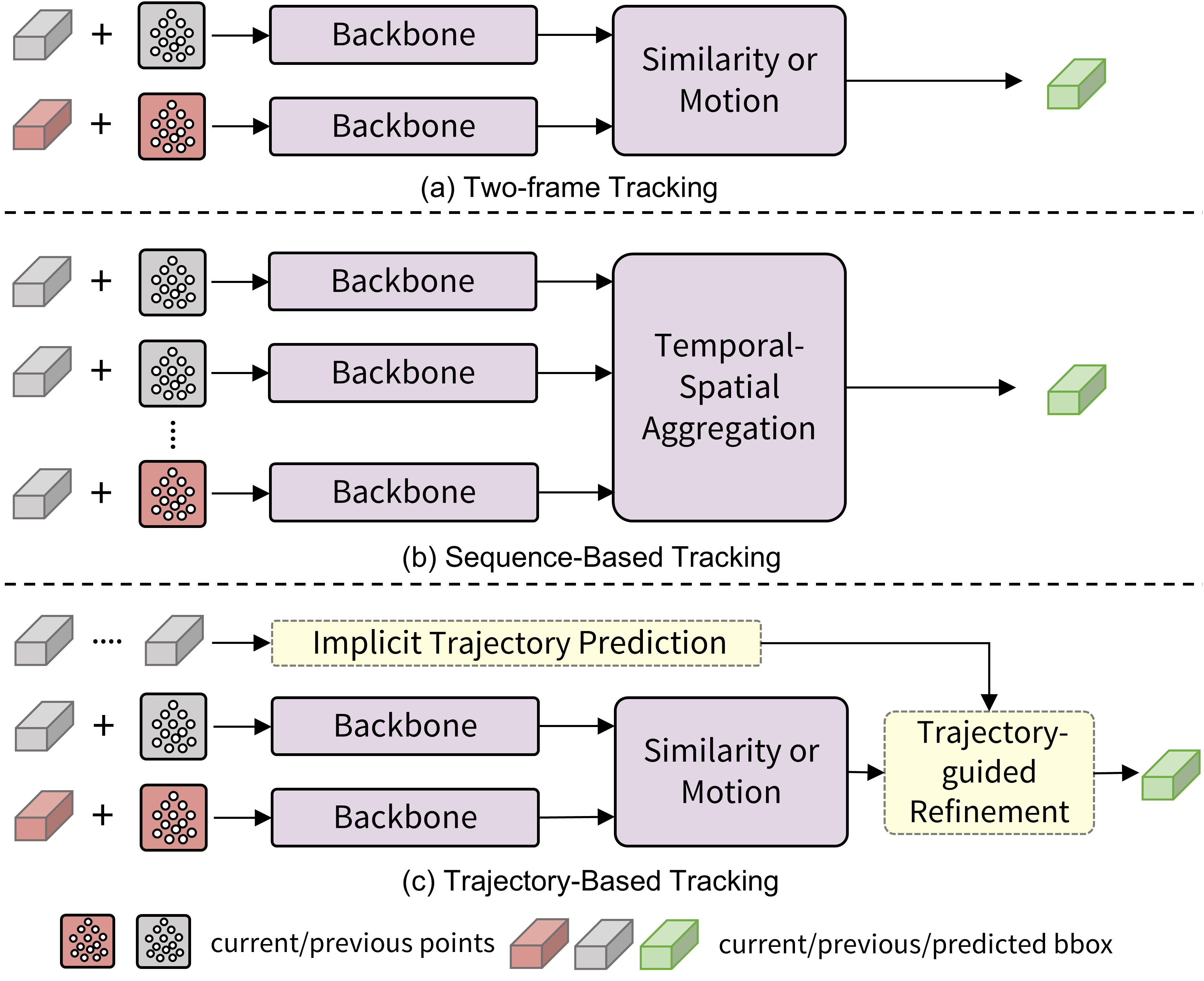}     \caption{\textbf{Different tracking paradigms.} (a) Two-frame paradigm exploits two-frame inputs for tracking through appearance matching or motion prediction. (b) Sequence-based paradigm uses multi-frames inputs to integrate target information. (c) Our Trajectory-based paradigm considers both short- and long-term motion clues.} 
    \label{fig:tracking_paradigms}
    \vspace{-6mm}
\end{figure}

Previous methods typically follow a two-frame paradigm (Fig.\ref{fig:tracking_paradigms}(a)), which can be categorized into appearance-based~\cite{sc3d,p2b,fang20203d,bat,ptt2021shan, ptt-journal,zhou2022pttr} and motion-based approaches~\cite{mmtrack,m2track++,li2024flowtrack,focustrack, zhou2025comptrack, xu2024pillartrack}. Similarity-based methods such as SC3D \cite{sc3d}, P2B \cite{p2b}, and PTT \cite{ptt2021shan} rely on frame-wise appearance cues between template and search point cloud and perform similarity matching, but they are sensitive to appearance changes and fast motion. In contrast, motion-based methods like $\mathrm{M}^2$-Track series~\cite{mmtrack, m2track++} and P2P \cite{p2p} explicitly model inter-frame motion to estimate relative translation. However, by relying solely on two-frame information, they all struggle in sparse or occluded scenes due to limited appearance or motion clues. More critically, they lack considerations of long-term motion continuity, preventing them from forming a predictive motion prior—for example, anticipating a vehicle's position as it moves through a temporary occlusion—which is essential for robust tracking.

To overcome these challenges, recent studies have explored the sequence-based paradigm (Fig.\ref{fig:tracking_paradigms}(b)), which incorporates long-term temporal information by processing multiple point cloud frames. For instance, STTracker~\cite{STT} processes a sequence of BEV feature maps and uses a deformable attention mechanism to aggregate spatio-temporal features. Going a step further, SeqTrack3D~\cite{seqtrack3d} utilizes sequences of both point clouds and their corresponding historical bounding boxes, using a Seq2Seq model to explicitly learn motion patterns. While these approaches improve robustness in sparse scenes, they introduce a critical trade-off. Their reliance on processing multi-frame point clouds inevitably leads to a high computational cost, making them less suitable for latency-sensitive, real-time tracking applications. Besides, their complex feature extraction from sequential data can still struggle to learn a clear, consistent motion trajectory, especially with noisy or occluded frames. This leaves a clear opening for a method that can leverage long-term motion continuity in a more lightweight and direct manner.

In this paper, we propose \textbf{TrajTrack}, a novel 3D SOT framework (Fig.\ref{fig:tracking_paradigms}(c)), that synergistically combines the strengths of short-term explicit motion and long-term implicit continuity through a two-stage process. \textbf{Stage 1: Explicit Motion Proposal:} We first employ an efficient, two-frame explicit motion model. By analyzing two inter-frame point clouds, this stage rapidly generates an initial tracking proposal. This proposal effectively captures instantaneous motion but can be prone to errors in sparse or occluded scenes. \textbf{Stage 2: Implicit Trajectory Prediction:} This is the core innovation of our framework. We introduce an implicit motion modeling module that operates solely on the historical sequence of bounding box. It uses a lightweight Transformer to learn the object's long-term motion continuity and predict future trajectory with confidence. \textbf{Post-process: Trajectory-guided Proposal Refinement:} Finally, we design a proposal refinement mechanism. It leverages the predicted trajectory from Stage 2, which embodies a long-term motion prior, to intelligently calibrate and correct the potentially inaccurate initial proposal from Stage 1. By integrating short-term explicit observations with long-term, consistent motion patterns, TrajTrack effectively captures both local and global motion cues. Its key innovation lies in enhancing tracking robustness by exploiting the intrinsic continuity of object motion, without the high computational cost of processing multiple point cloud frames. Extensive experiments demonstrate the superior performance and real-time speed of our approach. In summary, our main contributions are as follows:

\begin{itemize}

\item  \textbf{Trajectory-based Paradigm}: We propose a novel trajectory-based paradigm that leverages historical bboxes to incorporate long-term motion continuity, enhancing robustness without the overhead of multi-frame input.

\item \textbf{TrajTrack with Implicit Motion Modeling}: We instantiate this paradigm in TrajTrack, a framework featuring a novel Implicit Motion Modeling (IMM) module to provide predictive priors to synergize long-term continuity with short-term observations.

\item  \textbf{SOTA Performance}: We achieve new state-of-the-art performance on the large-scale nuScenes benchmark by a significant margin (+4.48\% in Precision). Besides, we demonstrate the strong generalizability by consistently improving the existing method's performance.

\end{itemize}

%% file: text/2_related.tex
\vspace{-4mm}
\section{Related Work}
\vspace{-1mm}

\noindent\textbf{Object Tracking on Point Clouds.} 
The field of 3D single-object tracking (SOT) has advanced rapidly with the development of point cloud processing. The pioneering work SC3D~\cite{sc3d} applied Siamese networks and template matching directly to point clouds, but suffered from inefficiency and non-end-to-end training. To address this, P2B~\cite{p2b} and 3D-SiamRPN~\cite{fang20203d} employed RPN and VoteNet~\cite{votenet} for efficient proposal generation. Subsequent methods enhanced feature representation: BAT~\cite{bat} incorporated box-aware structural features, PTT~\cite{ptt2021shan,ptt-journal} leveraged transformers, and V2B~\cite{v2b} converted sparse points to dense BEV maps. Attention mechanisms were further integrated by STNet~\cite{hui2022stnet}, LTTR~\cite{lttr}, and PTTR~\cite{zhou2022pttr} to propagate target-specific features.
Appearance-based SOTs remain vulnerable under sparsity and occlusion, motivating motion-centric designs. $\mathrm{M}^2$-Track~\cite{mmtrack} modeled inter-frame motion via a segmentation-and-tracking pipeline, extended to semi-supervised settings by M²Track++~\cite{m2track++}. DMT~\cite{dmt} leveraged historical bounding boxes for center prediction, while P2P~\cite{p2p} inferred relative motion directly from cropped point clouds of consecutive frames. However, these methods primarily rely on short-term temporal inputs, neglecting long-term history. To address this, TAT~\cite{tat} aggregated historical templates via RNNs but struggled with low-quality inputs. STTracker~\cite{STT} employed deformable attention to integrate spatio-temporal features from BEV sequences, and SeqTrack3D~\cite{seqtrack3d} proposed a Sequence-to-Sequence paradigm to capture both geometry and motion from point cloud sequences. While sequence-based approaches improve robustness, they struggle to balance accuracy and efficiency due to the high cost of multi-frame processing and the complexity of modeling consistent motion. 

Similarly, 3D MOT exploits temporal cues but along different lines. Filter-based methods (AB3DMOT~\cite{AB3DSOT}, SimpleTrack~\cite{simpletrack}) use Kalman or constant-velocity models implicitly maintain motion through state vectors but struggle with nonlinearity. Query-based tracking-by-attention methods (MUTR3D~\cite{mutr3d}, ADA-Track~\cite{ada}), propagate object identities across frames via attention yet remain detection-driven and focus on association rather than explicit motion modeling. Crucially, MOT methods reinitialize states from detector outputs each frame; by contrast, SOT bypasses per-frame detection through initial target specification, enabling detector-free, persistent, and end-to-end tracking.

\noindent\textbf{Sequence Modeling for Motion.}
Learning motion patterns from temporal sequences provides an effective way to model long-term context. In trajectory prediction, VectorNet~\cite{vectornet} and LaneRCNN~\cite{LaneRCNN} introduced vectorized and rasterized scene representations with graph neural networks to capture dynamic interactions. GATraj~\cite{cheng2023gatraj} further enhanced this through graph attention. TNT~\cite{tnt} and DenseTNT~\cite{densetnt} improved long-term reasoning via hierarchical goal selection and dense trajectory generation. Generative approaches such as VAEs~\cite{wu2024smart}, CVAEs~\cite{trajectron}, and GANs~\cite{gupta2018sgan} learned trajectory distributions in a probabilistic manner. Most of these methods adopt a sequence-to-sequence paradigm, predicting future motion from past trajectory features. Recently, Transformer-based architectures have become dominant due to their ability to model complex temporal dependencies, as demonstrated by S2TNet~\cite{s2tnet} for spatio-temporal interactions and AgentFormer~\cite{yuan2021agent} for agent-centric motion coherence, with CASPFormer~\cite{CASPFormer} introducing causal spatial priors for socially compliant forecasting. The success of these models in learning rich motion patterns from sparse positional sequences directly informs our approach. We are inspired from this powerful sequence modeling paradigm for the specific task of 3D SOT, designing a lightweight IMM module to learn motion continuity from an object's past trajectory.

%% file: text/3_method.tex
\vspace{-1mm}
\section{Method}
\vspace{-2mm}

\begin{figure*}[t]
    \centering
    \vspace{-15pt}
     \includegraphics[width=\linewidth]{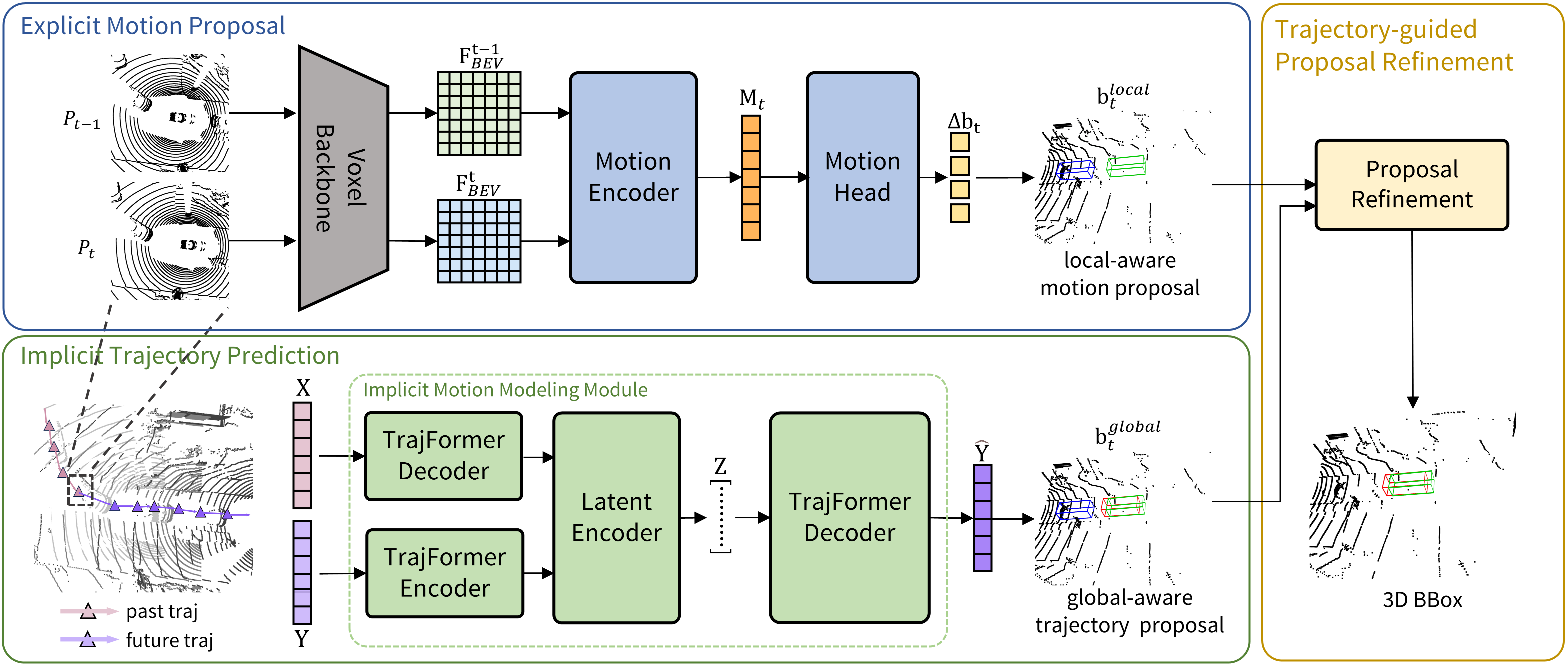}
    \caption{Overview of the proposed \textbf{Trajectory-Based Paradigm} and its instantiation \textbf{TrajTrack}. Explicit Motion Proposal uses a two-frame tracking baseline to obtain the local-aware motion proposal. Implicit Trajectory Prediction is used to learn the object’s global-aware motion continuity for trajectory proposal. Finally, Trajectory-guided Proposal Refinement cooperates the local and global-aware motion cues to get the refined 3D BBOX as the final output.}
    \label{fig:model}
\vspace{-4.5mm} 
\end{figure*}

\subsection{Preliminaries} 
\noindent\textbf{3D Single Object Tracking.}
In the 3D SOT task, given a sequence of point clouds $\mathbf{P}_{t}\in \mathbb{R}^{N\times3}$ at time t with N points and the initial 3D bounding box $
\mathbf{b}_t=(x_t,y_t,z_t,h_t,w_t,l_t,\theta_t)\in \mathbb{R} ^{1\times 7}
$ in the initial frame t, where $(x,y,z)$ and $(h,w,l)$ represent the center coordinate and size and $ \theta$ represent the rotation angle. The goal of 3D SOT aims to predict the box $
\mathbf{b}_{s}=(x_s,y_s,z_s,h_s,w_s,l_s, \theta_s)\in \mathbb{R} ^{1\times 7} $
in subsequent frames from the search area point cloud. Generally, the object size $(h,w,l)$ will not change in all frames; therefore, the position parameters $(x_s,y_s,z_s,\theta_s)$ need to be predicted.

\vspace{-2mm}
\subsection{Framework Overview} 

Our \textbf{TrajTrack} framework is designed to address the challenges of 3D SOT by synergistically combining short-term and long-term motion cues. As illustrated in Fig.~\ref{fig:model}, it follows a two-stage "propose-predict-refine" pipeline: \textbf{(1) Explicit Motion Proposal:} First, a two-frame-based explicit motion model rapidly generates an initial, locally-aware tracking proposal for the current frame. \textbf{(2) Implicit Trajectory Prediction:} Second, our novel Implicit Motion Modeling (IMM) module predicts a globally-aware future trajectory. \textbf{(3) Trajectory-Guided Proposal Refinement:} Finally, a post-processing operation from the refinement mechanism uses the trajectory prior to intelligently correct the initial proposal, outputting a more robust tracking result.

\vspace{-3mm}
\subsection{Stage 1: Explicit Motion Proposal} 
This stage aims to rapidly generate a high-quality initial proposal that captures instantaneous motion. Recent works~\cite{p2p, lu2024voxeltrack} have demonstrated the superiority of voxels as a 3D representation in single object tracking (SOT). Therefore, we employ an efficient, voxel-based backbone from ~\cite{p2p} to extract BEV features $\mathbf{F}_{t-1}, \mathbf{F}_t \in \mathbb{R}^{H\times W}$ from consecutive point clouds $\mathbf{P}_{t-1}$ and $\mathbf{P}_t$. Notably, we can also use other 3D Tracker~\cite{p2b,mmtrack} in this stage; we also provide the details in Tab.~\ref{tab:Different Baselines}. These feature maps are concatenated and passed through a motion encoder and a motion head to predict the inter-frame relative motion $\Delta \mathbf{b}_t$:
\begin{align}
\mathbf{M}t &= \mathrm{ME}([\mathbf{F}^{t-1}{BEV}, \mathbf{F}^t_{BEV}]) \\
\Delta \mathbf{b}_t &= \mathrm{MH}(\mathbf{M}t)
\end{align}
where $\mathrm{ME}$ and $\mathrm{MH}$ are motion encoder and motion head, respectively. Finally, the initial motion proposal $\mathbf{b}_t^{\text{local}}$ for the current frame is obtained by applying this displacement to the previous frame's bounding box $\mathbf{b}{t-1}$:
\begin{align}
\mathbf{b}_t^{\text{local}} = \mathbf{b}_{t-1} + \Delta \mathbf{b}_t
\end{align}
This local-aware proposal effectively captures local motion but can be prone to errors in sparse or occluded scenes. More details about the Motion Encoder and head can refer to \cite{p2p}.

\begin{figure*}
    \centering
     \vspace{-17pt}
    \includegraphics[width=0.95\linewidth ]{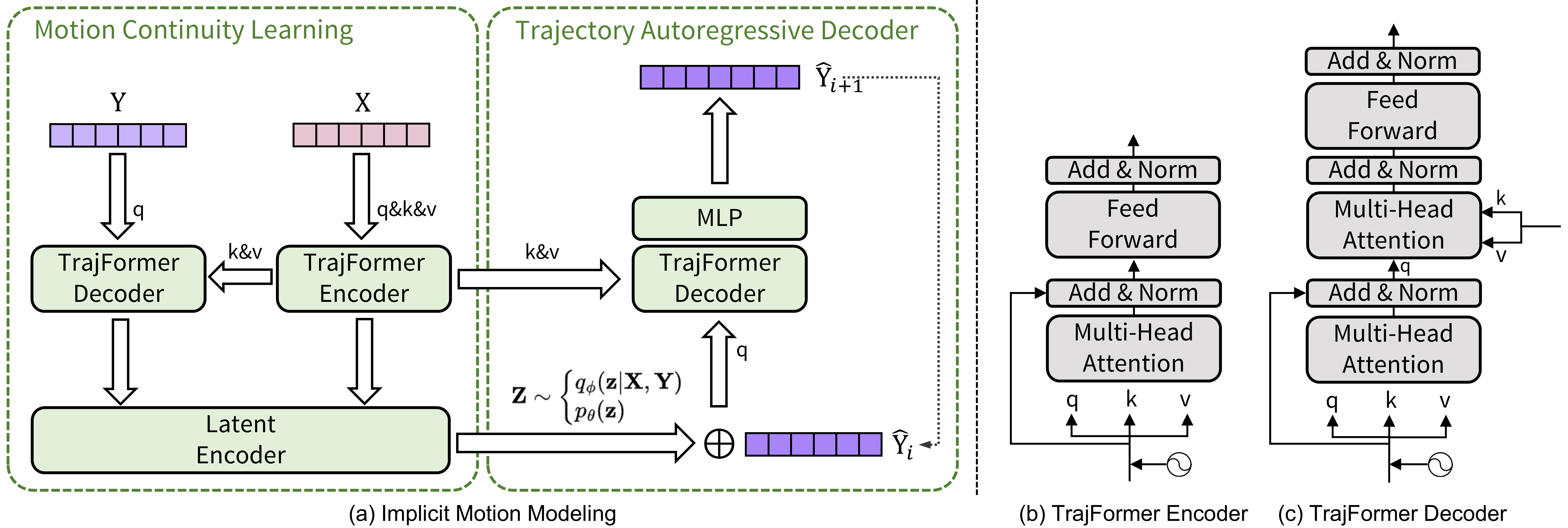}
    \caption{\textbf{(a) Framework of Implicit Trajectory Prediction. (b) The TrajFormer Encoder. (c) The TrajFormer Decoder.}}
    \label{fig:IMM}
\vspace{-5mm}
\end{figure*}

\vspace{-2mm}
\subsection{Stage 2: Implicit Trajectory Prediction} 
The core of our framework is the Implicit Motion Modeling (IMM) module, which learns long-term motion continuity to predict a robust future trajectory. Critically, this module operates solely on the lightweight historical sequence of past bounding box coordinates, $\mathbf{X}=(x^{-H+1},x^{-H+2},\ldots,x^{0})$, avoiding the high cost of processing multiple point clouds. The IMM module is trained to predict the future trajectory $\mathbf{Y}=(y^{0},y^{1},\ldots,y^{T})$ conditioned on the past trajectory $\mathbf{X}$ and a latent variable $\mathbf{Z}$ that captures the inherent stochasticity of motion. $H$ and $T$ denote historical/future timesteps, respectively. We utilize the input information to predict a set of future trajectories $\hat{\mathbf{Y}}$, from which the trajectory position at the corresponding time step is selected as the global-aware trajectory proposal $\mathbf{b}_t^{global}$. The IMM consists of two main components, which are both implemented using our proposed TrajFormer architecture (Fig.~\ref{fig:IMM}).

\noindent \textbf{Motion Continuity Learning}. This component learns a latent representation of motion dynamics. We derive motion-centric features from the observed trajectory $\mathbf{X}$ and encode them with TrajFormer encoder and use a MLP to  predict statistical parameters($\mu$, $\sigma$) to obtain prior distribution $p_{\theta}(\mathbf{Z}|\mathbf{X})=\mathcal{N}(\boldsymbol{\mu}^{p},\mathrm{Diag}(\boldsymbol{\sigma}^{p})^{2})$. During training, the TrajFormer decoder integrates the future trajectory $\mathbf{Y}$ as conditional input and fuses it with the observed history to predict the posterior distribution $q_{\phi}(\mathbf{Z}|\mathbf{Y},\mathbf{X})=\mathcal{N}(\boldsymbol{\mu}^{q},\mathrm{Diag}(\boldsymbol{\sigma}^{q})^{2})$. A KL divergence loss enforces consistency between the prior and posterior, enabling the latent variable to capture expressive motion patterns.

\noindent \textbf{Trajectory Autoregressive Decoder}. 
The prediction process is initiated by sampling a latent code $\mathbf{z}$ rom the appropriate distribution (the prior $p_{\theta}$ at inference, the posterior $q_{\phi}$ at training). This code $\mathbf{z}$, which encapsulates the global motion intent, is concatenated with the last observed state embedding $\mathbf{x}^0$ to form an initial feature vector $\mathbf{f}^0 = \mathbf{x}^0 \oplus \mathbf{Z}$. This initial vector initiates the Trajectory Autoregressive Decoder (TAD) through recurrent updates:
\begin{equation}
\mathbf{f}_t = \mathbf{y}_t \oplus \mathbf{Z}, \quad \mathbf{Z} \sim
\begin{cases}
q_{\phi}(\mathbf{Z}|\mathbf{X}, \mathbf{Y}) & \textit{(training)} \\
p_{\theta}(\mathbf{Z}|\mathbf{X}) & \textit{(inference)}
\end{cases}
\end{equation}
where $\mathbf{F}_\mathbf{Y}^t = \{\mathbf{f}^0,...,\mathbf{f}^t\}$ represents the progressively augmented feature sequence. The decoding mechanism maintains persistent latent guidance by keeping the sampled $\mathbf{Z}$ constant throughout generation, ensuring global motion pattern consistency. Simultaneously, it accumulates dynamic temporal context through the expanding feature sequence $\mathbf{F}_t$, which explicitly preserves historical state dependencies. At each timestep $t$, the decoder predicts the next trajectory feature $\mathbf{F}_{t+1}$ conditioned on this dual-information representation and then transformed through an MLP to reconstruct the predicted trajectory $\mathbf{y}_{t+1}$. Finally, we get the predicted future trajectory $\hat{\mathbf{Y}}$ of length $t_{f}$. The predicted position for the current timestep serves as our final global-aware trajectory proposal, $\mathbf{b}_t^{global}$.

\noindent \textbf{TrajFormer Architecture}. 
Both the encoder and decoder are built upon our TrajFormer block, which adapts the standard Transformer architecture for trajectory modeling. To provide temporal context, we first add a cosine-based positional encoding scheme to the input trajectory coordinates. A TrajFormer block consists of two main sub-layers: multi-head self-attention (MHSA) and a feed-forward network (FFN). For an input sequence $\mathbf{X}_{in}$, the $\mathbf{Q}$, $\mathbf{K}$, $\mathbf{V}$ matrices are first projected:
\begin{align}
\mathbf{Q} &= (\mathbf{X} + \mathrm{PE}(t_p))\mathbf{W}_{Q}, \\
\mathbf{K} &= (\mathbf{X} + \mathrm{PE}(t_p))\mathbf{W}_{K}, \\
\mathbf{V} &= (\mathbf{X} + \mathrm{PE}(t_p))\mathbf{W}_{V}
\end{align}

The self-attention output is then calculated, followed by residual connections, layer normalization, and an FFN block:
\begin{align}
\mathbf{X}_{\text{attn}} &= \mathrm{Softmax}\left(\frac{\mathbf{Q}\mathbf{K}^\top}{\sqrt{d_k}}\right)\mathbf{V}, \\
\mathbf{X}' &= \mathrm{LayerNorm}(\mathbf{X}_{in} + \mathbf{X}_{\text{attn}}) , \\
\mathbf{X}_{\text{out}} &= \mathrm{LayerNorm}(\mathbf{X}' + \mathrm{FFN}(\mathbf{X}'))
\end{align}
The TrajFormer Decoder block is similar but includes an additional cross-attention layer to incorporate the encoded features from the past trajectory.

\vspace{1mm}
\noindent \textbf{Insight and Advantages of IMM}. 
The core insight behind the IMM is the decoupling of long-term motion modeling from the high-bandwidth point cloud data. Unlike sequence-based methods that are burdened by processing multiple, dense point clouds to extract motion cues, our IMM operates on a highly compressed, low-dimensional representation: the historical trajectory of bounding box centers. This design is motivated by the observation that for tracking, the object's macro-level motion continuity (where it's going) is often more critical for overcoming occlusions and sparsity than its micro-level surface details in every historical frame. By learning from the trajectory alone, the IMM can focus exclusively on capturing the object's dynamic behavior over time—such as velocity and turning patterns—creating a robust, global motion prior. This lightweight, information-centric approach is the key to how \textbf{TrajTrack} achieves the robustness of a sequence-based method while only introducing a minor computational burden compared to a two-frame tracker.

\vspace{-2mm}
\subsection{Trajectory-guided Proposal Refinement}
The final post-process of TrajTrack is to dynamically fuse the proposals from the explicit and implicit motion models. This stage receives two inputs: the agile but potentially noisy local-aware proposal, $\mathbf{b}_t^{\text{local}}$, and the stable, global-aware trajectory proposal, $\mathbf{b}_t^{\text{global}}$. To synergize these two, we introduce a confidence-based refinement strategy. We use the Intersection-over-Union (IoU) between the two proposals themselves, IoU($\mathbf{b}_t^{\text{local}}$, $\mathbf{b}_t^{\text{global}}$), as a metric for the reliability of the short-term, explicit prediction. \textbf{(1)} If the IoU is high (above a threshold ${\lambda}_{IoU}$), it indicates that both short-term and long-term models are in agreement. In this case, we trust the more precise, local-aware proposal, $\mathbf{b}_t^{\text{local}}$, as the final output. \textbf{(2)} If the IoU is low, it signals a potential failure of the explicit model, likely due to sparsity or occlusion. In this scenario, we rely on the more stable, long-term trajecory proposal,  $\mathbf{b}_t^{\text{global}}$, as a robust fallback. This mechanism allows our tracker to be fast and precise in simple scenarios while leveraging the long-term motion prior to enhance robustness and recover from failures in challenging environments.

\begin{table*}[t]
\centering
\vspace{-15pt}
\caption{Comparison with SOTA methods on the nuScenes dataset. Success and Precision are used for evaluation, \textbf{Bold} and \underline{underline} denote the best result and the second-best one respectively. * 64159 indicates the number of instances of cars}
\resizebox{\linewidth}{!}{
\begin{tabular}{c|c|c|cccccc}
\toprule
Method&  Publish&Paradigm&Car
[64,159]*& Pedestrian
[33,227]& Truck
[13,587]& Trailer
[3,352]& Bus
[2,953]& Mean 
[117,278]\\
\midrule
\midrule
SC3D~\cite{sc3d}{}& CVPR'19 &  \multirow{12}{*}{Two-frame}&22.31 / 21.93 & 11.29 / 12.65 & 35.28 / 28.12 & 35.28 / 28.12 & 29.35 / 24.08 & 20.70 / 20.20\\
P2B~\cite{p2b} & CVPR'20 &  &38.81 / 43.81 & 28.39 / 52.24 & 48.96 / 40.05 & 48.96 / 40.05 & 32.95 / 27.41 & 36.48 / 45.08 \\
PTT~\cite{ptt-journal}& IROS'21 &  &41.22 / 45.26 & 19.33 / 32.03 & 50.23 / 48.56 & 51.70 / 46.50 & 39.40 / 36.70 & 36.33 / 41.72 \\
BAT~\cite{bat} & ICCV'21 &  &40.73 / 43.29 & 28.83 / 53.32 & 52.59 / 44.89 & 52.59 / 44.89 & 35.44 / 28.01 & 38.10 / 45.71 \\
V2B~\cite{v2b} & NIPS'21 & & 54.40 / 59.70 & 30.10 / 55.40 & 53.70 / 54.50 & 54.90 / 51.44 & - / - & - / - \\
M$^2$-Track~\cite{mmtrack}&CVPR'22  & &55.85 / 65.09 & 32.10 / 60.92 & 57.36 / 59.54 & 57.61 / 58.26 & 51.39 / 51.44 & 49.23 / 62.73 \\
PTTR~\cite{zhou2022pttr}&CVPR'22 &  &51.89 / 58.61 & 29.90 / 45.09 & 45.30 / 44.74 & 45.87 / 38.36 & 43.14 / 37.74 & 44.50 / 52.07 \\
GLT-T~\cite{nie2023glt}&AAAI'23 &  &48.52 / 54.29 & 31.74 / 56.49 & 52.74 / 51.43 & 57.60 / 52.01 & 44.55 / 40.69 & 44.42 / 54.33 \\
 MLSET~\cite{MLSET}& RAL'23& & 53.20 / 58.30& 33.20 / 58.60& 54.30 / 52.50& 53.10 / 40.90& - / -&- / -\\
PTTR++~\cite{pttr_journal}&PAMI'24 &  &59.96 / 66.73& 32.49 / 50.50 & 59.85 / 61.20& 54.51 / 50.28& 53.98 / 51.22& 51.86 / 60.63
\\
FlowTrack~\cite{li2024flowtrack} & IROS'24 & & 60.29 / 71.07 & 37.60 / 67.64 & - / - & 55.39 / 62.70 & - / - & - / - \\ 
 P2P~\cite{p2p}&IJCV'25 &  &\underline{65.15 / 72.90}& \underline{46.43 / 75.08}& \underline{64.96 / 65.96}& \underline{70.46 / 66.86}& \underline{59.02 / 56.56}&\underline{59.84 / 72.13}\\
\midrule
STTracker~\cite{STT}& RAL'23 &  \multirow{2}{*}{Sequence}&56.11 / 69.07 & 37.58 / 68.36 & 54.29 / 60.71& 48.13 / 55.40& 36.31 / 36.07 & 49.66 / 66.77 \\
SeqTrack3D~\cite{seqtrack3d}&ICRA'24 & & 62.55 / 71.46& 39.94 / 68.57& 60.97 / 63.04& 68.37 / 61.76& 54.33 / 53.52&55.92 / 68.94\\
\midrule
 \rowcolor{gray!30} \textbf{TrajTrack}&  Ours & Trajectory& \textbf{68.02 / 75.87}& \textbf{48.32 / 78.78}& \textbf{67.19 / 68.44}& \textbf{70.70 / 68.02}& \textbf{61.16 / 57.67}&\textbf{62.25 / 75.15}\\
\emph{Improvement}&  & &\textcolor{mygreen1}{$\uparrow$ \textbf{2.87} / $\uparrow$ \textbf{2.97}}& \textcolor{mygreen1}{$\uparrow$ \textbf{1.89} / $\uparrow$ \textbf{3.70}}& \textcolor{mygreen1}{$\uparrow$ \textbf{2.23} / $\uparrow$ \textbf{2.48}}& \textcolor{mygreen1}{$\uparrow$ \textbf{0.24} / $\uparrow$ \textbf{1.16}}& \textcolor{mygreen1}{$\uparrow$ \textbf{2.14} / $\uparrow$ \textbf{1.11}}& \textcolor{mygreen1}{$\uparrow$ \textbf{2.41} / $\uparrow$ \textbf{3.02}}\\
\bottomrule
\end{tabular}
}
\label{tab:NuScenes}
\vspace{-4mm}
\end{table*}

\vspace{-4mm}
\subsection{Loss Function} 
Our \textbf{TrajTrack} framework is trained end-to-end with a composite loss function, $\mathcal{L}_{total}$, which jointly optimizes the explicit motion proposal and the implicit trajectory prediction.

\noindent \textbf{Explicit Motion Loss ($\mathcal{L}_{\mathrm{tracking}}$).}
To supervise the initial proposal from Stage 1, we follow~\cite{p2p} and adopt the Residual Log-likelihood Estimation (RLE) loss~\cite{rle}, which reduces the difficulty of regression by adding a hand-designed residual term distribution to the L2 loss term, making the fitted distribution closer to the true distribution, as the tracking loss:
\begin{align}
\mathcal{L}_{\mathrm{tracking}}=
\mathcal{L}_{\mathrm{RLE}}(\mathbf{b}_t^{\text{local}},\mathbf{b}_t^{gt})
\end{align}

\noindent \textbf{Implicit Trajectory Prediction Loss ($\mathcal{L}_{\mathrm{traj}}$).} To train our IMM module from Stage 2, we use the standard negative Evidence Lower Bound (ELBO) loss \cite{elbo}. This loss comprises a reconstruction term, which ensures the predicted trajectory matches the ground truth, and a KL-divergence term, which regularizes the latent space:
\begin{align}
\mathcal{L}_\mathrm{pred} = \mathcal{L}_{elbo} & =-\mathbb{E}_{q_{\phi}(\mathbf{Z}|\mathbf{Y},\mathbf{X})}[\log p_{\theta}(\mathbf{Y}|\mathbf{Z},\mathbf{X})] \\ 
& + \mathrm{KL}(q_{\phi}(\mathbf{Z}|\mathbf{Y},\mathbf{X})\|p_{\theta}(\mathbf{Z}|\mathbf{X}))
\end{align}

\noindent \textbf{Overall Objective.} The final training objective is a weighted sum of these two losses:
\begin{align}
\mathcal{L}_{total}=\mathcal{L}_{\mathrm{tracking}}+\lambda\mathcal{L}_{\mathrm{traj}}
\end{align}
where $\lambda$ hyper-parameter that balances the contribution of the trajectory prediction task.

%% file: text/4_exp.tex
\vspace{1mm}
\section{Experiments}
\noindent\textbf{Dataset and Metrics.} 
We follow the common setup~\cite{mmtrack,hu2025mvctrack} and conduct experiments on the large-scale nuScenes~\cite{nuScenes} dataset. Notably, due to the limited data size of the KITTI dataset (only 19 training, 2 validation sequences~\cite{ptt-journal,nie2023glt}) makes it challenging to adequately evaluate the methods. In contrast, the nuScenes dataset comprises 700 training and 150 validation sequences, across 40K point cloud frames, allowing for a more comprehensive evaluation. The evaluation metrics is followed the common setup~\cite{ptt2021shan,ptt-journal} to report \textit{Success} and \textit{Precision} based on one pass evaluation (OPE)~\cite{otb2013,kristan2016novel}.

\noindent\textbf{Implementation Details.}
We follow previous works~\cite{p2b,p2p,ptt-journal}, set the previous frame $t{-}1$ is set as the template and the current frame $t$ as the search region. These cropped regions, defined as 
$[(-4.8, 4.8), (-4.8, 4.8), (-1.5, 1.5)]$ for cars and 
$[(-1.92, 1.92), (-1.92, 1.92), (-1.5, 1.5)]$ for humans along the $(x, y, z)$ axes. Additionally, the data augmentation, such as random horizontal flipping and uniform rotation within [-5$^\circ$,5$^\circ$] is applied to target points and bounding boxes to improve the model’s generalization. The parameters of IMM are set to $H=2, T=12, {\lambda}_{IoU}=0.5$. All experiments are conducted on NVIDIA RTX 3090 GPUs. The network is trained for 20 epochs with the AdamW optimizer, initialized with a learning rate of 0.0001, with a batch size is 32. Our experiment video is available at \textcolor{blue}{\url{https://www.bilibili.com/video/BV1ahYgzmEWP}}.

\vspace{-3mm}
\subsection{Comparison on nuScenes dataset:}
To rigorously evaluate our framework under challenging real-world conditions, we conduct comprehensive comparisons on the challenging nuScenes dataset, characterized by complex scenes and sparser point clouds from 32-beam LiDARs data in diverse scenes. Following the standard protocol for SOT evaluation, we compare exclusively against state-of-the-art SOT methods. As shown in Tab.~\ref{tab:NuScenes}, our TrajTrack consistently outperforms all prior trackers across all categories, demonstrating its exceptional scalability and robustness in sparse and diverse environments. For instance, TrajTrack it surpasses a strong baseline, P2P~\cite{p2p}, by 2.87/ 2.97\% (Success/Precision) in the Car category and by an even larger margin of 1.89/3.70\% in the Pedestrian category. This robust performance in a sparse environment highlights the core advantage of our trajectory-based paradigm: by leveraging lightweight, long-term motion continuity, our method maintains a stable track where methods relying solely on instantaneous cues falter. 
\vspace{-2mm}
\begin{table}[h]
\Large
\caption{Comparison of the running speeds on different representative methods.}\label{table:speed}
\resizebox{\linewidth}{!}{

\begin{tabular}{c|cccc}
\toprule[.05cm]
Method & STTracker~\cite{STT} & LTTR\cite{lttr} & GLT-T~\cite{nie2023glt} & SeqTrack3D~\cite{seqtrack3d}\\ 
FPS    & 22.0 & 23.0 & 30.0 & 38.0\\ 
\midrule
Method & P2B\cite{p2b} & PTT\cite{ptt-journal} & PTTR++~\cite{pttr_journal} & M$^2$Track\cite{mmtrack}\\ 
FPS    & 40.0 & 40.0 & 43.0 & 51.2\\ 
\midrule
Method  & \cellcolor{gray!20} TrajTrack (Ours) & BAT\cite{bat} & M$^2$Track++\cite{m2track++} & \textbf{P2P}~\cite{p2p} \\ 
 FPS   & \cellcolor{gray!20} 54.7 & 57.0 & 57.0 & \textbf{63.9}\\ 
\bottomrule[.05cm]
\end{tabular}
}
\vspace{-1mm}
\end{table}

\noindent\textbf{Running Speed:}
Inference speed is also a vital factor for practical applications. We present a comprehensive speed comparison of TrajTrack with other methods in Tab.~\ref{table:speed}. Following common evaluation protocols~\cite{p2b, ptt2021shan,p2p}, speed is measured by calculating the average running time of all frames in the Car category. On a single NVIDIA RTX 3090 GPU, TrajTrack achieves 54.7 FPS. Despite the computational overhead incurred by our IMM module, TrajTrack yields significant performance improvements, maintaining a better trade-off between accuracy and speed.

\begin{table}[ht]
\centering
\vspace{-2mm}
\caption{Ablation Study on Different Baselines}
\resizebox{0.7\linewidth}{!}{
\begin{tabular}{lcc}
\toprule
Method & $\textbf{M}^\textbf{2}$\textbf{-Track}\cite{mmtrack}& \textbf{BAT}\cite{bat}\\
\midrule
Baseline & 55.85 / 65.09 & 40.73 / 43.29 \\
w/ TrajTrack   & 58.16 / 68.23 & 47.07 / 51.66 \\
\midrule
\emph{improvement}& $\uparrow$ \textcolor{mygreen1}{\textbf{2.31 / 3.14}}& $\uparrow$ \textcolor{mygreen1}{\textbf{6.34 / 8.37}}\\
\bottomrule
\end{tabular}
}
\label{tab:Different Baselines}
\vspace{-2mm}
\end{table}

\noindent\textbf{Generalizability Across Different Trackers:} To demonstrate the versatility of our trajectory-based paradigm, we replaced two distinct baselines into our framework with Implicit Motion Modeling (IMM) module. As illustrated in the Tab.~\ref{tab:Different Baselines}, when applied our method to the similarity-based paradigm BAT\cite{bat} and the motion-based paradigm $\mathrm{M}^2$-Track\cite{mmtrack}, consistently and significantly achieving improvements of 2.31/3.14\% and 6.34/8.37\% in Success/Precision, respectively.This demonstrates that our approach of leveraging long-term motion continuity is a general principle that can enhance various 3D SOT architectures, regardless of their underlying paradigm.

\begin{figure}[h]
    \vspace{-2mm}
    \centering
    \includegraphics[width=0.85\linewidth]{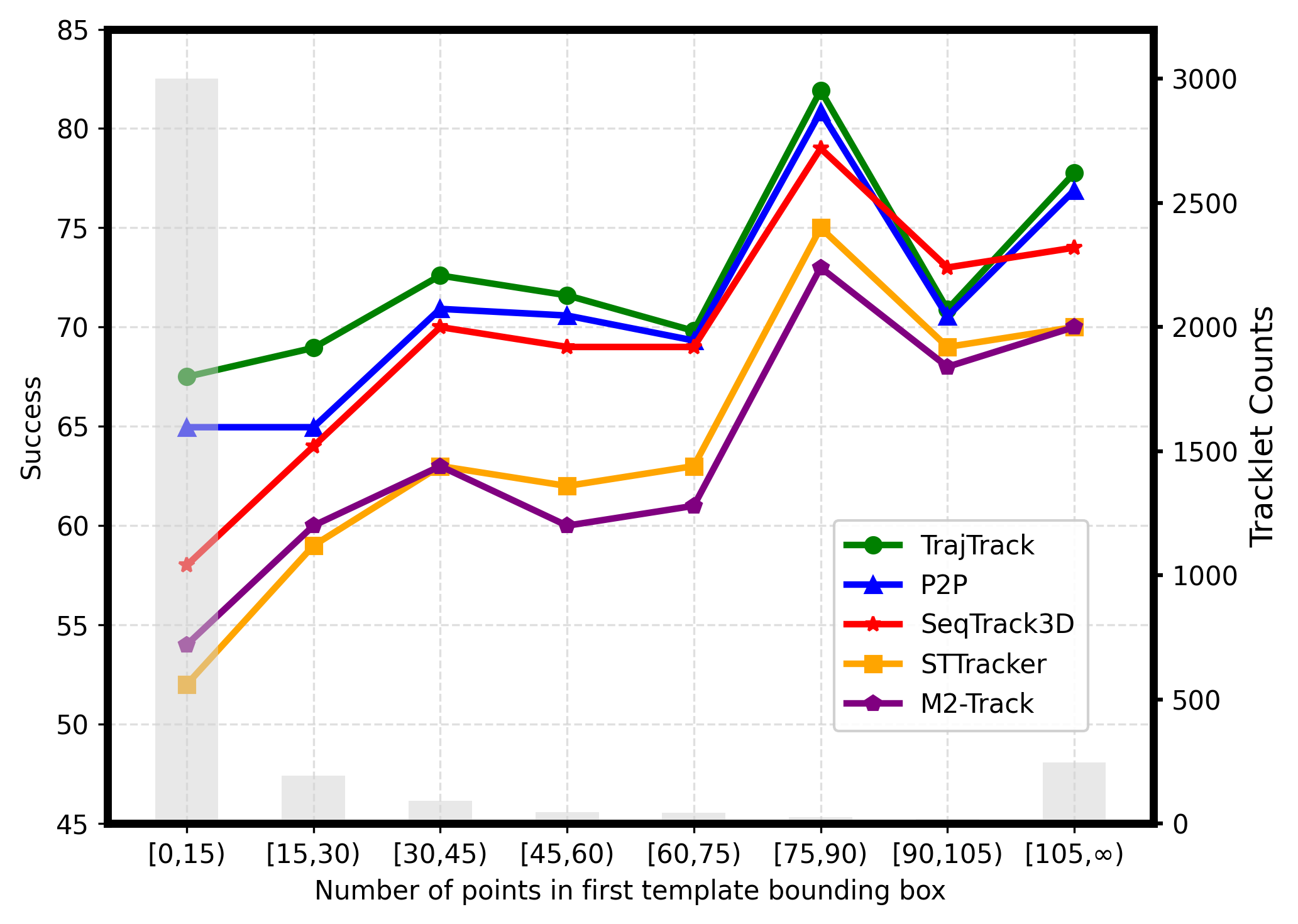}
    \vspace{-2mm}
    \caption{\textbf{Tracking performance across varying numbers of template points in the first frame.}}
    \label{fig:different points tracklet}
    \vspace{-2mm}
\end{figure}

\noindent\textbf{Robustness to Sparsity:} We evaluated our method's performance under varying levels of point cloud sparsity by analyzing its accuracy based on the number of points in the initial template. As shown in Fig.~\ref{fig:different points tracklet}, TrajTrack consistently outperforms other leading methods, especially in extremely sparse scenarios. On the nuScenes dataset, where a majority of tracklets begin with fewer than 15 points, our method's advantage is most pronounced. This highlights its superior ability to leverage long-term motion continuity when instantaneous appearance cues are minimal. To further validate this, we identified a challenging subset where over half the frames contain fewer than 20 points. On these 3,061 tracklets, our method achieves a Success/Precision of 67.08\% / 75.39\%, significantly outperforming the baseline's 65.16\% / 72.57\%, which confirms its enhanced robustness in low-density conditions.
\vspace{-2mm}
\begin{table}[h]
\centering
\caption{Ablation Study of components in TrajTrack.}
\label{tab:components}
\resizebox{0.8\linewidth}{!}{
 \vspace{-6mm}
\begin{tabular}{c|cc|cc}
\hline
 \textbf{IMM Setting}& \multicolumn{2}{c|}{\textbf{Car}} & \multicolumn{2}{c}{\textbf{Ped}} \\
    & Succ. & Prec. & Succ. & Prec. \\
\hline
Baseline~\cite{p2p}   & 65.15& 72.90& 46.43& 75.08\\
MLP-based IMM  & 67.27& 74.14& 47.68& 77.33\\
 \rowcolor{gray!20}  TrajFormer-based IMM &
\textbf{68.02}& \textbf{75.87}& \textbf{48.32}& \textbf{78.78}\\
 \midrule
\end{tabular}
}
\vspace{-3mm}
\end{table}

\vspace{-3mm}
\subsection{Ablation Study}
\vspace{-1mm}

\noindent\textbf{Ablation of IMM:}  We analyze the core Implicit Motion Modeling module of our method in Tab.~ \ref{tab:components}. Compared to the baseline tracker~\cite{p2p}, which relies solely on explicit two-frame motion, incorporating a simple MLP-based IMM already yields substantial performance gains, especially for the Pedestrian category (+2.25\% Success). This result validates the core premise of our work: leveraging implicit motion continuity from past trajectories is highly effective. By further employing our more powerful TrajFormer-based IMM, the performance is enhanced to its peak. This systematically demonstrates that while the IMM concept itself is beneficial, the TrajFormer architecture is key to fully capturing the complex temporal dependencies in the trajectory data.

 \vspace{-1mm}
\begin{table}[htbp]
\centering
\caption{Ablation Study on Different timesteps Values}
\label{tab:timesteps}
\resizebox{0.7\linewidth}{!}{
\begin{tabular}{ccc}
\toprule
timesteps& Precision (\%) & Success (\%) \\
\midrule
$H=1$(Baseline)& 65.15& 72.90\\
$H=2, T=12$& 68.02& \textbf{75.87}\\$H=3, T=6$& 67.17& 75.29\\
 $H=3, T=12$& \textbf{68.07}&75.81\\
$H=4, T=12$& 67.63& 75.67\\
\bottomrule
\end{tabular}
}
\end{table}
 \vspace{-2mm} 
\begin{table}[htbp]
\centering
\caption{Ablation Study on Different ${\lambda}_{IoU}$ Values}
\label{tab:lambda_ablation}
\resizebox{0.7\linewidth}{!}{
\begin{tabular}{cccc}
\toprule
$\lambda$ & Precision (\%) & Success (\%) & Latency (ms) \\
\midrule
0.20 & 66.15& 74.26& 17.77 \\
0.30 & 66.81& 74.83& 17.98 \\
0.40 & 67.46& 75.32& 18.06 \\
0.50 & 68.02& 75.87& 18.27 \\
\bottomrule
\end{tabular}
}
 \vspace{-1mm}
\end{table}
\noindent\textbf{Hyperparameters:} 
We conducted a hyperparameter ablation on the historical length $H$, prediction horizon $T$, and refinement threshold ${\lambda}_{IoU}$ (Tab. \ref{tab:timesteps}-\ref{tab:lambda_ablation}). The results show that $H = 2 $, $T = 12$ provide the best overall performance, achieving the highest Success score while maintaining competitive Precision. This configuration offers sufficient temporal context without introducing redundant history. In addition, ${\lambda}_{IoU} = 0.5$  yields the highest accuracy with only marginal computational overhead, and is therefore adopted as the default setting.

\begin{figure}[t]
    \centering
    \vspace{-10pt}
    \includegraphics[width=\linewidth]{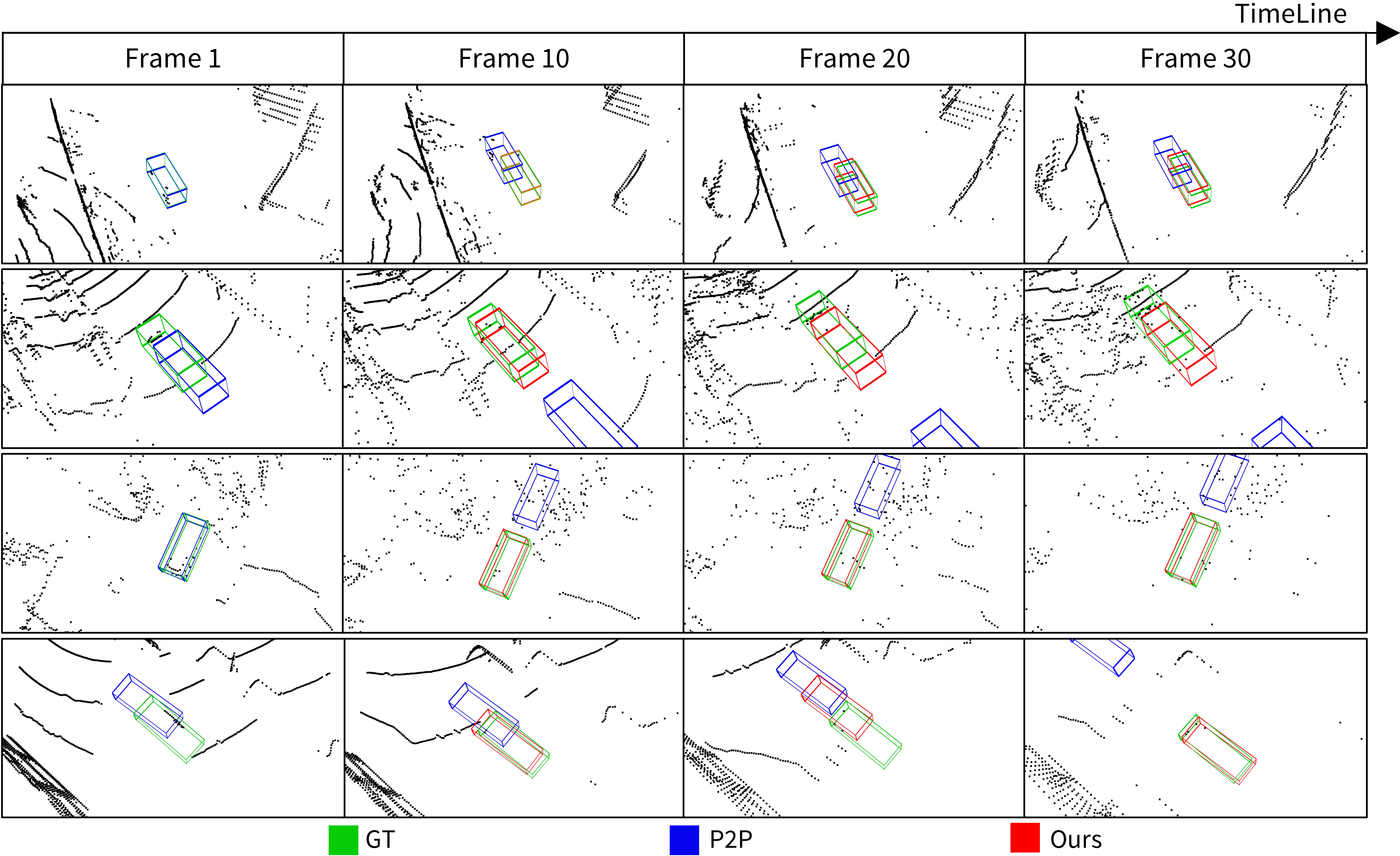}
    \caption{\textbf{Visualization results on nuScenes.} Starting from the second frame, the global-aware trajectory proposal from Implicit Motion Modeling corrects the biased tracking results, achieving accurate tracking. }
    \label{fig:visualization}
    \vspace{-5mm}
\end{figure}
\begin{figure}[!h]
    \centering
    \includegraphics[width=\linewidth]{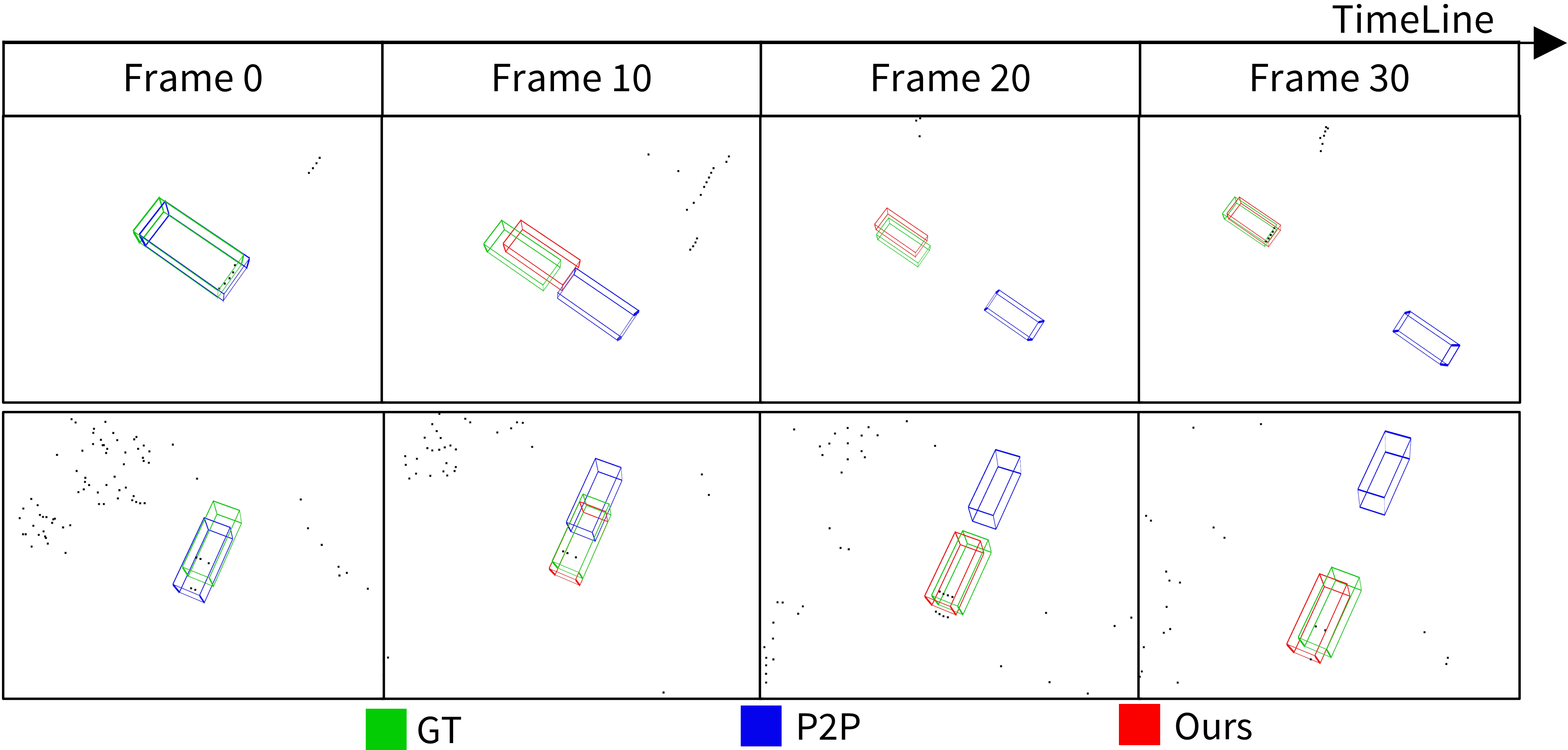}
    \caption{\textbf{Visualization results on sparse point cloud scenes.}}
    \label{fig:visualization-sparse}
\vspace{-10pt}
\end{figure}
\vspace{-2mm}
\begin{figure}[!h]
    \centering
    \includegraphics[width=0.95\linewidth]{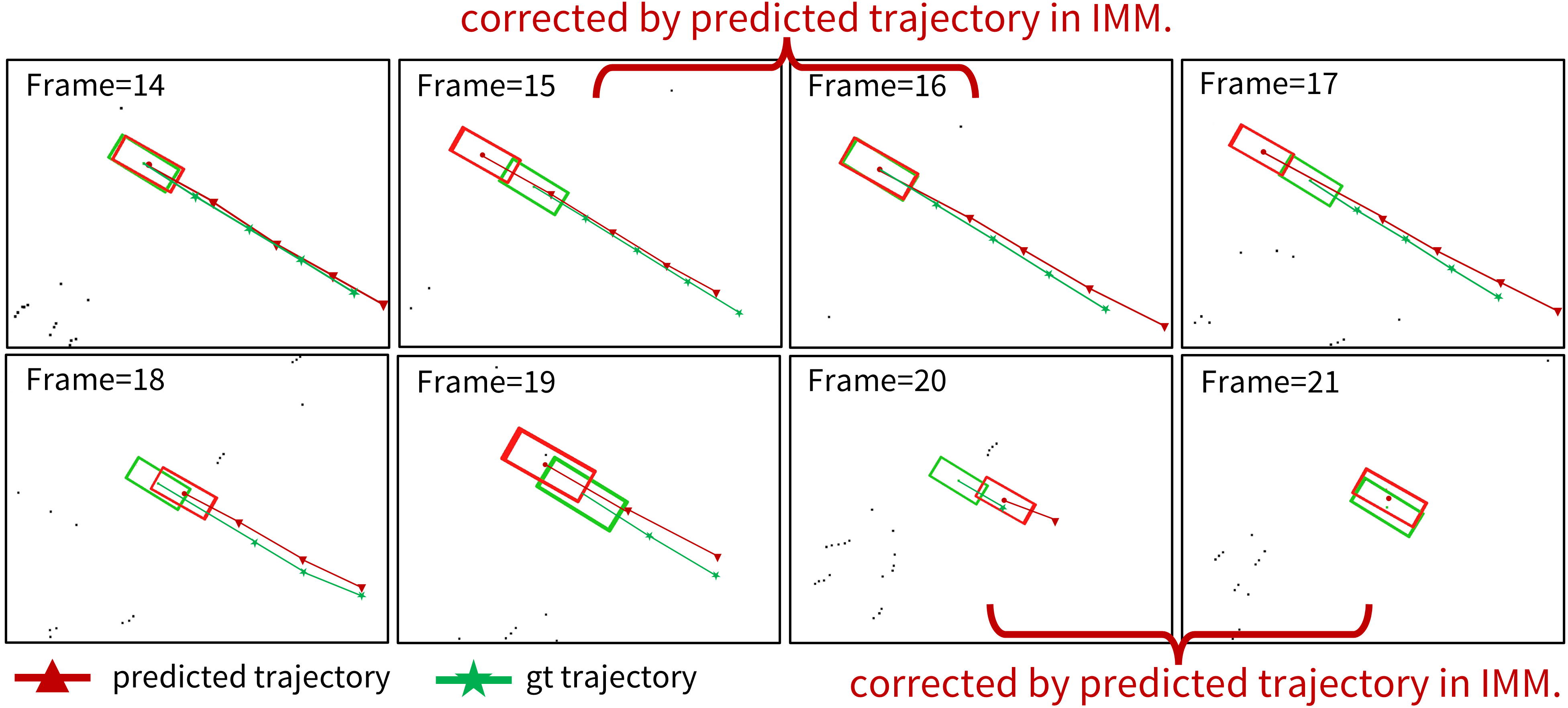}
    \caption{\textbf{Visualization results with Implicit Trajectory Prediction correction.}}
    \label{fig:visualization_correct}
    \vspace{-6mm}
\end{figure}

\subsection{Visualization results.}
To provide a qualitative evaluation, we visualize our tracking results against a strong baseline~\cite{p2p} in various challenging scenes. As shown in Fig.~\ref{fig:visualization}, our TrajTrack consistently demonstrates more accurate and stable tracking than the baseline. When the baseline tracker begins to fail due to occlusion or sparsity, our Trajectory-guided Proposal Refinement mechanism integrates a global-aware trajectory proposal from the Implicit Motion Modeling. This allows our method to leverage the long-term motion prior to correct misaligned bounding boxes and effectively recover the track, preventing the failure from error accumulation.Additionally, as illustrated in Fig. \ref{fig:visualization_correct}, while tracking failed, the subsequent prediction can still be corrected by leveraging long-term motion information. Furthermore, Fig.~\ref{fig:visualization-sparse} highlights our method's robustness in extremely sparse scenes. Even when the target is represented by only a few scattered points, TrajTrack successfully leverages temporal consistency to track accurately. These visualizations confirm that the synergistic integration of short-term motion and long-term trajectory priors significantly enhances tracking reliability in dynamic and complex scenes.

\vspace{-4mm}
\begin{figure}[!h]
    \centering
    \includegraphics[width=\linewidth]{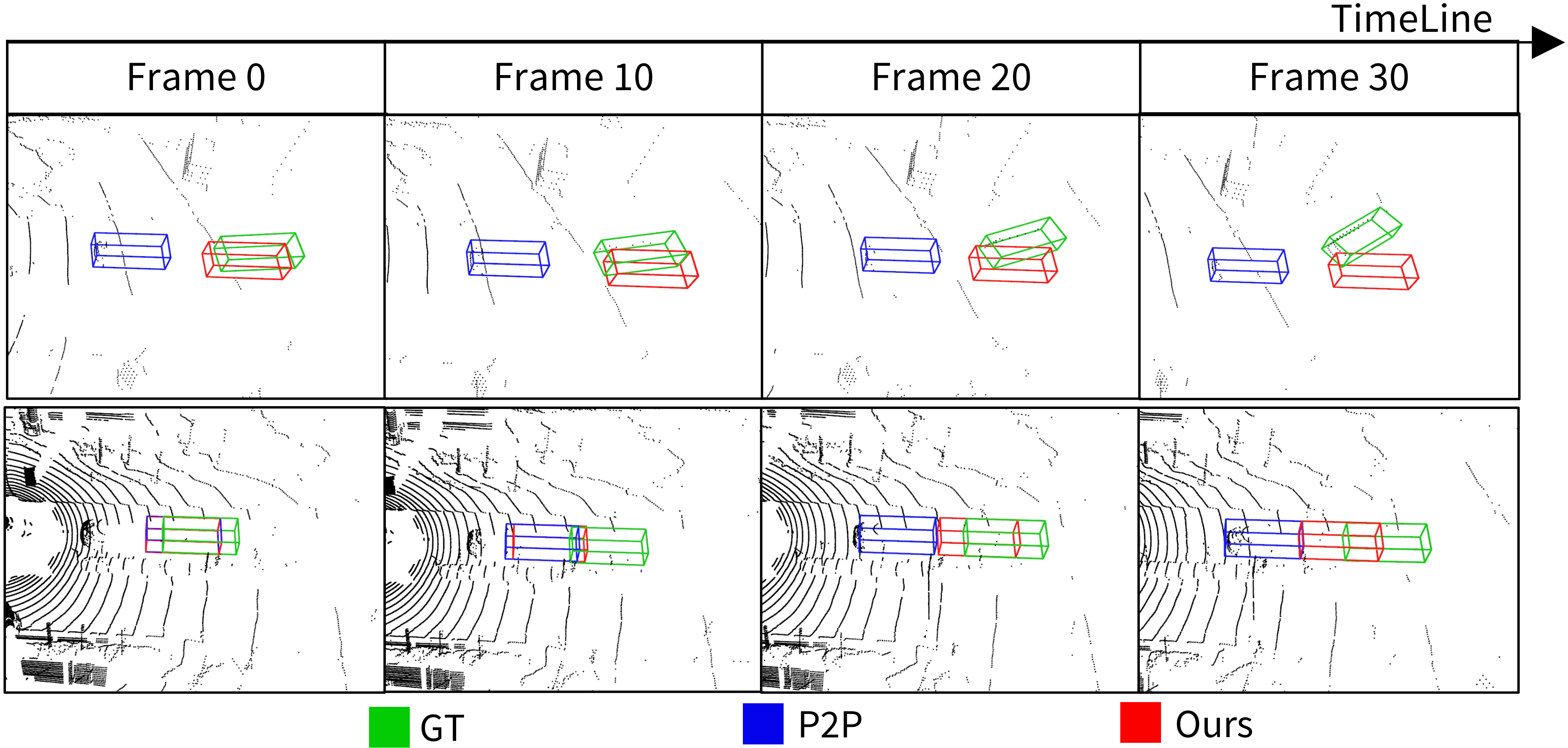}
    \caption{\textbf{Visualization results on failed tracklets.}}
    \label{fig:visualization-fail}
\vspace{-10pt}
\end{figure}

%% file: text/5_conclusion.tex
\section{CONCLUSIONS}

In this work, we introduced \textbf{TrajTrack}, a novel trajectory-based paradigm for 3D single object tracking that resolves the critical trade-off between the robustness of two-frame methods and the efficiency of sequence-based approaches. Our key innovation is an Implicit Motion Modeling (IMM) module that leverages lightweight, historical bounding box trajectories to learn long-term motion continuity. By synergistically refining a short-term, explicit motion proposal with this powerful long-term prior, TrajTrack achieves a new state-of-the-art tracking performance in the challenging nuScenes dataset, while maintaining real-time speed. For future work, we plan to explore tighter fusion strategies between the explicit and implicit modules and investigate the application of this lightweight temporal modeling approach to other robotics perception tasks. Beyond this, future extensions could augment the current trajectory-only prior with lightweight structured auxiliary cues, such as ego-motion or semantic priors, as broader predictive modeling studies suggest that informative non-traditional structured signals can improve forecasting quality~\cite{liu2026health}. Another promising direction is explainable trajectory modeling, where feature-attribution tools may help reveal which historical states or contextual variables dominate the predicted motion prior, improving model diagnosis and transparency~\cite{shen2025aienhanced}.

\noindent\textbf{Limitations:} Despite outperforming baselines in highly non-linear scenarios (e.g., sharp turns, Fig. \ref{fig:visualization-fail}), TrajTrack's reliance on motion continuity limits full convergence to the ground truth in these extreme cases. This indicates a need for more adaptive motion modeling to handle abrupt kinematic transitions. In addition, recent evidence from challenging predictive settings suggests that robustness can depend not only on architecture but also on objective design and operating-point selection~\cite{sun2025objective}, which motivates future exploration of risk-aware losses and failure-oriented optimization for abrupt or rare motion patterns. Furthermore, replacing the current heuristic proposal refinement with a learnable mechanism could yield further improvements. Besides, we can adopt quantization methods~\cite{zhoulidarptq,jiang2025ptq4ris,wang2025point4bit,xumambaquant,hu2025ostquant,yu2025mquant,yu2025fq} to accelerate 3D Tracker for resource-limited application.

\vspace{-2mm}